\documentclass{isprs} 
\usepackage{subfigure}
\usepackage{setspace}
\usepackage{geometry} 
\usepackage{epstopdf}
\usepackage[labelsep=period]{caption}  
\usepackage[british]{babel}
\usepackage[hang]{footmisc}

\usepackage{graphicx} 
\usepackage{amsmath}
\usepackage{amssymb}

\usepackage[table]{xcolor}  
\usepackage{colortbl}

\usepackage{svg}
\usepackage{booktabs}
\usepackage{adjustbox}
\usepackage{textcomp}
\usepackage{url}
\usepackage{multirow}
\usepackage{makecell}
\usepackage{wrapfig}
\usepackage{float}
\usepackage{dblfloatfix}

\definecolor{errOK}{HTML}{1B9E77}      
\definecolor{errWarn}{HTML}{E69F00}    
\definecolor{errBad}{HTML}{D55E00}     

\begin{document}

\title{An RTK-SLAM Dataset for Absolute Accuracy Evaluation in GNSS-Degraded Environments}

\date{}

\author{Wei Zhang, Vincent Ress, David Skuddis, Uwe Soergel, Norbert Haala}

\address{Institute for Photogrammetry and Geoinformatics, University of Stuttgart, Germany -  firstname.lastname@ifp.uni-stuttgart.de}

\abstract{
RTK-SLAM systems integrate simultaneous localization and mapping (SLAM) with real-time kinematic (RTK) GNSS positioning, promising both relative consistency and globally referenced coordinates for efficient georeferenced surveying. A critical and underappreciated issue is that the standard evaluation metric, Absolute Trajectory Error (ATE), first fits an optimal rigid-body transformation between the estimated trajectory and reference before computing errors. This so-called SE(3) alignment absorbs global drift and systematic errors, making trajectories appear more accurate than they are in practice, and is unsuitable for evaluating the global accuracy of RTK-SLAM. We present a geodetically referenced dataset and evaluation methodology that expose this gap. A key design principle is that the RTK receiver is used solely as a \emph{system input}, while ground truth is established independently via a geodetic total station. This separation is absent from all existing datasets, where GNSS typically serves as (part of) the ground truth. The dataset is collected using a handheld RTK-SLAM device across two representative scenes, covering diverse GNSS conditions, including open-sky, building-obstructed areas, underpasses, outdoor-to-indoor transitions, and indoor environments. We evaluate LiDAR-inertial, visual-inertial, and LiDAR-visual-inertial RTK-SLAM systems alongside standalone RTK, reporting direct global accuracy and SE(3)-aligned relative accuracy to make the gap explicit. Results show that SE(3) alignment can underestimate absolute positioning error by up to 76\%. RTK-SLAM achieves centimeter-level absolute accuracy in open-sky conditions and maintains decimeter-level global accuracy indoors, where standalone RTK degrades to tens of meters. The dataset, calibration files, and evaluation scripts are publicly available at \url{https://rtk-slam-dataset.github.io/}.}


\maketitle

\section{Introduction}\label{MANUSCRIPT}

\sloppy
Accurate and reliable positioning is fundamental for surveying, geospatial data acquisition, and robotic navigation. Conventional geodetic techniques achieve this by establishing observations within global coordinate frames using GNSS receivers and total stations. While these methods can deliver millimeter-level accuracy, they are typically tied to static instrument setups or point-by-point observations with a survey pole, which makes data collection labor-intensive and time-consuming, especially in large or complex environments. Simultaneous localization and mapping (SLAM) has emerged as a complementary approach that provides accurate relative positioning by continuously estimating the trajectory of a sensor platform while reconstructing the surrounding space~\cite{Cadena2016}. SLAM operates effectively in GNSS-denied environments, such as urban canyons or indoor facilities, but its results remain confined to a local coordinate system. As a result, SLAM maps from different surveys cannot be directly integrated into geodetic or BIM coordinate frames without additional georeferencing effort.

Recent advances have led to the integration of SLAM with RTK GNSS positioning, often referred to as SLAM-RTK or RTK-SLAM systems~\cite{shan2020lio,wang2024givl}. These hybrid solutions combine the global accuracy of GNSS positioning with the robustness of SLAM, offering two key advantages. First, they provide globally referenced coordinates even when operating across outdoor and indoor domains, with SLAM constraining drift when GNSS signals are degraded or unavailable. Second, they enable mobile and efficient surveying: instead of occupying each checkpoint individually, surveyors can simply walk through the environment with a handheld device, continuously recording georeferenced data. This mobility lowers the barrier for large-scale and high-frequency surveys, making it attractive for construction monitoring, asset management, and rapid documentation tasks.

\begin{figure*}[h!]
\begin{center}
\includegraphics[width=\linewidth]{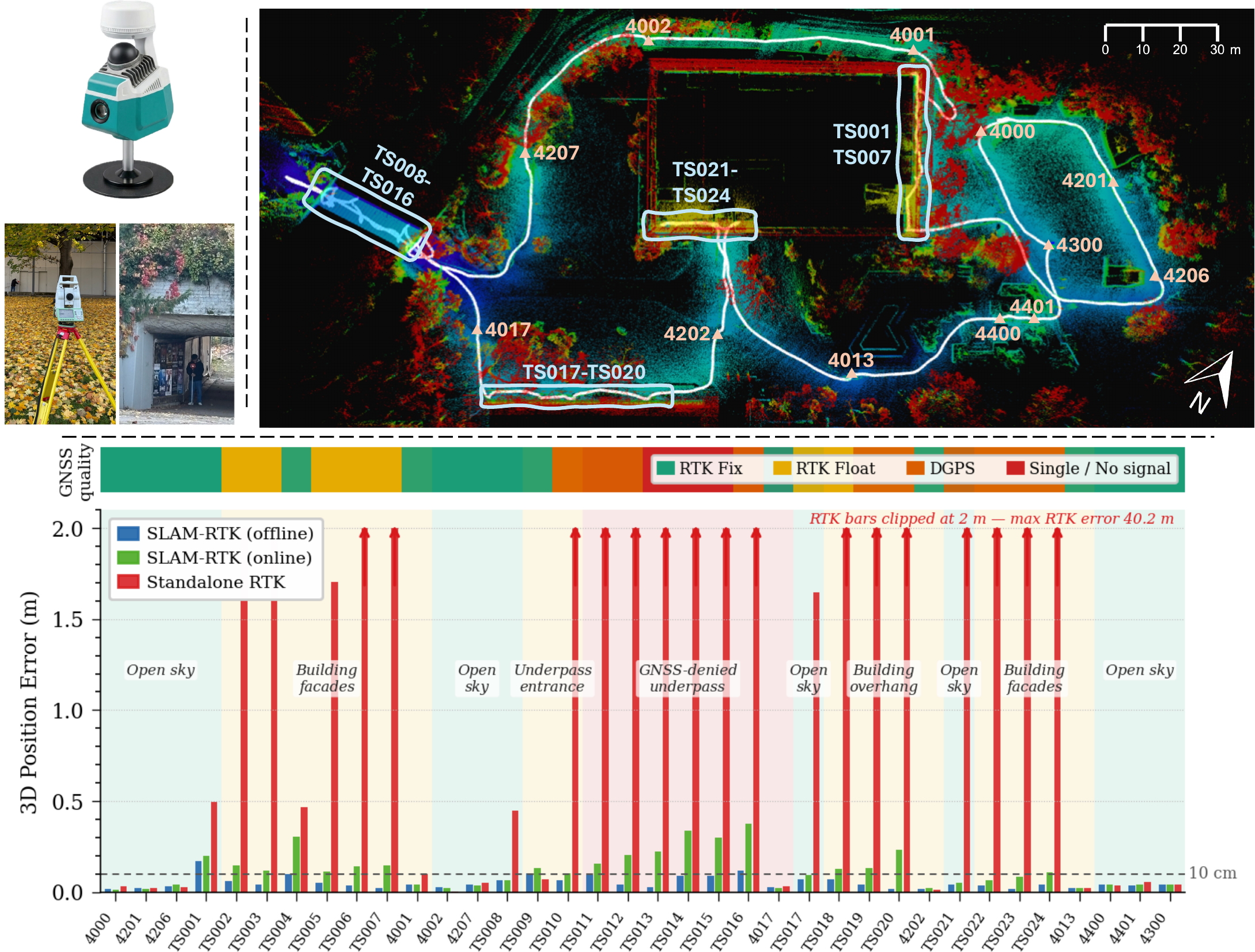}
\caption{
\textbf{Top:} Equipment setup (left) and overview of checkpoints overlaid on the SLAM map of the Stadtgarten scene (right).
Orange marked checkpoints are under open sky, while cyan-marked checkpoints are under GNSS obstruction (e.g. buildings, trees, underpass).
\textbf{Bottom:} Absolute 3D error per checkpoint for Stadtgarten Seq.\,1 using FAST-LIO-SAM method. Standalone RTK errors grow to tens of meters in GNSS-degraded zones, while offline RTK-SLAM remains mostly below 10\,cm.
}
\label{fig:teaser}
\end{center}
\end{figure*}

Despite this promise, important open questions remain. GNSS accuracy is strongly affected by multipath and obstruction~\cite{yuan2023multipath}, and the degree to which SLAM can maintain global accuracy when GNSS signals degrade or are lost has not been rigorously quantified. More importantly, existing SLAM benchmarks evaluate accuracy with SE(3)-aligned ATE~\cite{sturm2012benchmark}, which fits a rigid transformation between the estimated and reference trajectories before computing errors. This alignment is appropriate when only relative accuracy matters, but it is fundamentally incompatible with evaluating absolute global positioning (see Section~\ref{sec:eval_protocol} for a detailed discussion). A trajectory that is meters away from its true global position can still yield a near-zero SE(3)-aligned ATE if its internal geometry is consistent. For RTK-SLAM systems intended for georeferenced surveying, this distinction is critical.

To address these gaps, we propose a dedicated dataset and evaluation methodology. The contributions are:
\begin{itemize}
    \item \textbf{A geodetically referenced RTK-SLAM dataset} covering outdoor GNSS-degraded and indoor GNSS-denied environments, with RTK measurements used exclusively as system inputs, not as ground truth. Sub-centimeter ground truth is established independently via geodetic total station. To our knowledge, this is the first dataset enabling direct evaluation of absolute global RTK-SLAM accuracy across outdoor and indoor scenes with this separation.
    \item \textbf{An evaluation methodology} for absolute global accuracy without SE(3) alignment, applicable to any SLAM system that outputs globally referenced coordinates. Comparing against standard SE(3)-aligned ATE reveals that the alignment can underestimate absolute global errors by up to 76\,\%, masking critical failures.
    \item \textbf{Benchmarking of various RTK-SLAM systems} across scenes, reporting both online and offline results, with analysis of positioning drift in relation to GNSS outages. We show that while standalone RTK degrades to tens of meters, offline LiDAR-aided methods can maintain decimeter-level absolute accuracy in challenging GNSS-degraded conditions.
\end{itemize}

\section{Related Work}

\begin{table*}[htbp]
\centering
\small
\caption{Comparison of representative SLAM and positioning datasets.
    \textit{RTK role}: whether RTK is used as system input or ground truth.
    \textit{Abs.\ geodetic}: absolute accuracy assessable without SE(3) alignment.
    \checkmark\,=\,yes, --\,=\,no, $\sim$\,=\,partial.}
\label{tab:dataset_comparison}
\adjustbox{max width=\linewidth,center}{%
\setlength{\tabcolsep}{4.5pt}
\begin{tabular}{@{}lllllrcc@{}}
\toprule
\textbf{Dataset} & \textbf{Environment} & \textbf{Sensors} & \textbf{GT type} & \textbf{RTK role} & \textbf{GT accuracy} & \textbf{Abs.\ geodetic} \\
\midrule
\shortstack[l]{KITTI\\\cite{geiger2013vision}}           & Urban roads        & LiDAR, Cam, IMU, GNSS & RTK/INS               & GT      & $<$10\,cm            & $\sim$      \\
\shortstack[l]{EuRoC\\\cite{burri2016euroc}}             & Indoor lab         & Cam, IMU              & Motion capture        & --      & $<$1\,mm             & --          \\
\shortstack[l]{MulRan\\\cite{kim2020mulran}}             & Urban outdoor      & LiDAR, Radar, IMU, GNSS & RTK/INS             & GT      & $<$10\,cm            & $\sim$      \\
\shortstack[l]{OpenLORIS\\\cite{shi2020openloris}}       & Indoor scenes      & LiDAR, RGB-D, Cam, IMU  & Mocap / 2D LiDAR SLAM & --    & $<$1\,mm / 10\,cm    & --          \\
\shortstack[l]{Hilti-Oxford\\\cite{zhang2022hilti}}         & Construction       & LiDAR, Cam, IMU       & Laser scanner         & --      & $<$1\,mm             & --          \\
\shortstack[l]{WHU-Helmet\\\cite{li2023whuhelmet}}       & Campus out/indoor  & LiDAR, Cam, IMU, GNSS & FOG-IMU + PPK + LiDAR & GT      & cm-level             & $\sim$      \\
\shortstack[l]{MCD\\\cite{nguyen2024mcd}}                & Multi-campus       & LiDAR, Cam, IMU, UWB  & LiDAR + Survey map    & --      & cm-level             & --          \\
\shortstack[l]{M2DGR+\\\cite{yin2024m2dgr}}              & Campus out/indoor  & LiDAR, Cam, IMU, GNSS & Mocap / RTK           & GT      & $<$1\,mm / 2\,cm     & $\sim$      \\
\shortstack[l]{FPortV2\\\cite{wei2025fusionportablev2}}  & Campus + urban     & LiDAR, Cam, IMU, GNSS & TS + RTK              & GT      & 1\,mm / 2\,cm        & $\sim$      \\
\midrule
\textbf{Ours} & \textbf{Park + construction} & \textbf{LiDAR, Cam, IMU, RTK} & \textbf{TS + static GNSS} & \textbf{Input} & $<$\textbf{1\,cm} & \checkmark \\
\bottomrule
\end{tabular}}
\end{table*}
\subsection{LiDAR-Inertial and Visual-Inertial Odometry}
LiDAR-based SLAM has advanced significantly in recent years. LOAM~\cite{zhang2014loam} introduced edge and planar feature extraction for scan-to-map registration. LIO-SAM~\cite{shan2020lio} extended this with a factor graph backend supporting IMU pre-integration and GNSS factors, enabling globally consistent mapping. FAST-LIO2~\cite{xu2022fast} replaced feature extraction with direct point-to-map registration using an incremental k-d tree, achieving higher efficiency on solid-state LiDARs such as the Livox MID360. On the other hand, visual-inertial odometry (VIO) fuses camera and IMU measurements to estimate trajectory without LiDAR. Leutenegger~et~al.~\cite{leutenegger2015keyframe} pioneered keyframe-based VIO using nonlinear optimization, demonstrating that tightly coupled camera-IMU factor graphs yield accurate and consistent state estimates. VINS-Mono~\cite{qin2018vins} extended this with sliding-window optimization and loop closure for robust monocular operation. OKVIS2~\cite{leutenegger2022okvis2} scaled the keyframe-based formulation to multi-camera setups with real-time loop closure.

\subsection{SLAM with GNSS Integration}
Both LiDAR-inertial and visual-inertial systems can serve as odometry front-ends within a factor graph that additionally incorporates GNSS position factors, enabling globally referenced trajectory estimation. FAST-LIO-SAM~\cite{wang2022fast_lio_sam} exemplifies this for LiDAR-inertial systems, combining the FAST-LIO2 front-end with the LIO-SAM graph optimization backend. For visual-inertial systems, VINS-Fusion~\cite{qin2018vins,qin2025general} extends VINS-Mono with GNSS fusion and a general multi-sensor factor graph, enabling globally referenced trajectories. Lightweight systems combining GNSS with visual-inertial odometry for seamless indoor-outdoor navigation have also been proposed~\cite{angelats2021iopes}. OKVIS2-X~\cite{boche2025okvis2} takes this further with a configurable system supporting visual-inertial-GNSS (VIG) and LiDAR-visual-inertial-GNSS (LVIG) configurations within the same keyframe-based backend. Wang~et~al.~\cite{wang2024givl} explore tighter GNSS-inertial-visual-LiDAR coupling. Despite these advances, these existing methods have not been evaluated in terms of direct global accuracy against geodetic ground truth.

\subsection{SLAM Benchmarks and Datasets}
A number of multi-sensor SLAM datasets have been proposed for benchmarking odometry and mapping systems. KITTI~\cite{geiger2013vision} provides a large-scale outdoor benchmark, while EuRoC~\cite{burri2016euroc} targets visual–inertial odometry in indoor environments. The Hilti SLAM Challenge~\cite{zhang2022hilti} focuses on construction scenarios with handheld devices but evaluates only relative accuracy. MulRan~\cite{kim2020mulran} offers long-term outdoor sequences without geodetic ground truth. Recent datasets such as MCD~\cite{nguyen2024mcd}, FusionPortableV2~\cite{wei2025fusionportablev2}, M2DGR+~\cite{yin2024m2dgr}, and WHU-Helmet~\cite{li2023whuhelmet} expand sensor diversity and environments, yet none systematically evaluate absolute global accuracy of RTK-SLAM across outdoor and indoor scenes against independent geodetic references. Geodetic evaluation using total stations has been applied to UWB-based indoor positioning~\cite{dipietra2020mobile}, but not to mobile SLAM across mixed environments. Similarly, indoor mapping benchmarks~\cite{wang2020isprs} and handheld LiDAR SLAM comparisons~\cite{trybala2023handheld} lack globally referenced evaluation. Most benchmarks rely on SE(3)-aligned ATE~\cite{sturm2012benchmark}, which is unsuitable for assessing global accuracy. A key limitation is the role of GNSS: in existing datasets (e.g., KITTI, MulRan, M2DGR+, FusionPortableV2, WHU-Helmet), GNSS serves as (part of) the ground truth rather than an input modality, preventing meaningful assessment of how well GNSS-aided SLAM maintains absolute accuracy. In contrast, our dataset uses RTK strictly as system input, while ground truth is established independently via total station and static GNSS observations. Table~\ref{tab:dataset_comparison} summarizes these distinctions across representative datasets.

\section{Dataset}

This section describes the dataset collected for this work. We first introduce the sensor platform and its calibration, then describe the two data collection scenes and the geodetic ground truth establishment procedure.

\subsection{Sensor Platform}
Data acquisition was performed using a handheld RTK-SLAM device comprising a Livox MID360 LiDAR sensor with integrated IMU, a 2\,megapixel global shutter camera, and a UM980 GNSS receiver. The Livox MID360 provides a $360^\circ$ horizontal and $59^\circ$ vertical field of view with a non-repetitive scan pattern, operating at 10\,Hz. IMU measurements are recorded at 200\,Hz. Differential GNSS corrections were provided by the German SAPOS service~\cite{riecken2017satellitenpositionierungsdienst}, enabling the GNSS receiver to operate in RTK mode and achieve centimeter-level positioning accuracy under open-sky conditions.
\begin{figure}[h!]
\centering
\includegraphics[width=0.60\columnwidth]{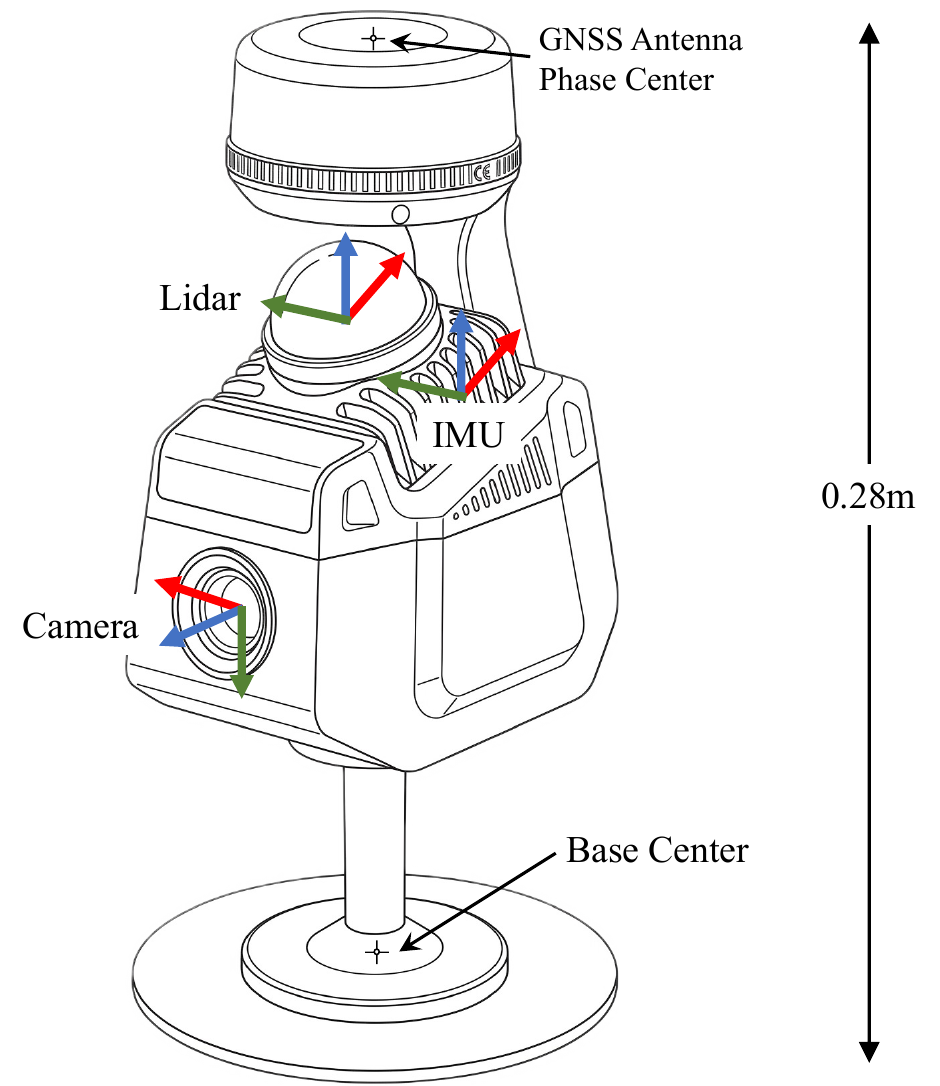}
\caption{Sensor platform with coordinate frames (red\,=\,$x$, green\,=\,$y$, blue\,=\,$z$) for LiDAR, IMU, and camera. GNSS antenna phase center (top) references RTK measurements. The base center (bottom) is matched against checkpoints.}
\label{fig:scanner_sketch}
\end{figure}

Figure~\ref{fig:scanner_sketch} illustrates the physical layout of the device and the individual coordinate frames of each sensor. The GNSS antenna phase center at the top of the device is the reference point for RTK position measurements, while the base center at the bottom of the pole is the physical reference point matched against surveyed checkpoints. All inter-sensor rigid body transformations are expressed relative to the IMU frame. The dataset is released in ROS bag and EuRoC format. To comply with privacy requirements, all camera images were anonymized prior to release by blurring human faces using \textit{deface}~\cite{deface2026}.

\subsection{Sensor Calibration}
\begin{table}[htbp]
\centering
\small
\caption{Sensor calibration parameters.}
\label{tab:calibration}
\adjustbox{max width=\columnwidth,center}{%
\begin{tabular}{@{}lll@{}}
\toprule
\textbf{Parameter} & \textbf{Value} & \textbf{Method} \\
\midrule
\multicolumn{3}{@{}l}{\textit{Camera intrinsics (1600$\,\times\,$1200\,px)}} \\
$f_x,\;f_y$ (px)   & 890.60,\enspace 890.62              & OpenCV checkerboard \\
$c_x,\;c_y$ (px)   & 780.97,\enspace 589.79              & OpenCV checkerboard \\
$k_1,\;k_2$        & $-$0.1471,\enspace 0.0810           & OpenCV checkerboard \\
$p_1,\;p_2$        & $-3.94{\times}10^{-4}$,\enspace $6.56{\times}10^{-4}$ & OpenCV checkerboard \\
\midrule
\multicolumn{3}{@{}l}{\textit{Camera $\to$ IMU extrinsic $\mathbf{T}_{C\to I}$}} \\
Rotation (r/p/y, deg) & $0.21,\enspace 67.44,\enspace -89.84$   & Kalibr \\
Translation (m)        & $(0.022,\enspace -0.048,\enspace -0.058)$ & Kalibr \\
Time offset (s)        & $-0.0206$ (camera delay)            & Kalibr \\
\midrule
\multicolumn{3}{@{}l}{\textit{LiDAR $\to$ IMU extrinsic $\mathbf{T}_{L\to I}$}} \\
Rotation               & identity                             & Livox MID360 manual \\
Translation (m)        & $(-0.011,\enspace -0.023,\enspace 0.044)$ & Livox MID360 manual \\
\midrule
\multicolumn{3}{@{}l}{\textit{Reference point offsets from IMU (m)}} \\
GNSS antenna phase center & $(0.023,\enspace -0.023,\enspace 0.090)$ & CAD model \\
Base center               & $(-0.073,\enspace -0.023,\enspace -0.172)$ & CAD model \\
\bottomrule
\end{tabular}}
\end{table}
Accurate extrinsic calibration and precise time synchronization are critical prerequisites for high-accuracy multi-sensor fusion, as uncompensated spatial offsets or temporal misalignment directly degrade positioning accuracy. Table~\ref{tab:calibration} summarizes the intrinsic and extrinsic parameters for all sensors. Camera intrinsics were obtained using the checkerboard method~\cite{zhang2000flexible} in OpenCV with a $400\,\text{mm}\times600\,\text{mm}$ target. Camera-to-IMU extrinsics and the temporal offset were jointly estimated using Kalibr~\cite{furgale2013unified} with the same target and sufficient IMU excitation. LiDAR-to-IMU extrinsics are adopted from the device user manual. The offsets of the GNSS antenna phase center and the base center relative to the IMU frame were derived from the CAD model. The LiDAR and its built-in IMU are hardware-synchronized to GNSS time via a 1\,PPS signal from the GNSS receiver. The camera is triggered by the SoC with a $-20.6\,\text{ms}$ delay relative to the IMU, as estimated by Kalibr.

\begin{table}[htbp]
\centering
\small
\caption{Dataset overview. The dataset comprises two scenes with four sequences in total.}
\label{tab:dataset}
\adjustbox{max width=\columnwidth,center}{%
\begin{tabular}{@{}lrrrr r@{}}
\toprule
\textbf{Sequence} & \textbf{Duration} & \textbf{Length} & \textbf{RTK Fix[\%]} & \textbf{Ctrl.\ Pts} & \textbf{Type} \\
\midrule
Stadtgarten \,1 & 26\,min\,42\,s & 1.04\,km & 54 & 36 & Outdoor park \\
Stadtgarten \,2 & 14\,min\,36\,s & 0.46\,km & 40 & 19 & Outdoor park \\
\midrule
Constr.\ Hall\,1 & 12\,min\,21\,s & 0.48\,km & 25 & 16 & Out.+Indoor \\
Constr.\ Hall.\,2 & \phantom{0}9\,min\,59\,s & 0.39\,km & 23 & 16 & Out.+Indoor \\
\bottomrule
\end{tabular}}
\end{table}

\begin{figure}[htbp]
\centering
\begin{tabular}{@{}c@{\hspace{2pt}}c@{}}
    \includegraphics[width=0.49\linewidth]{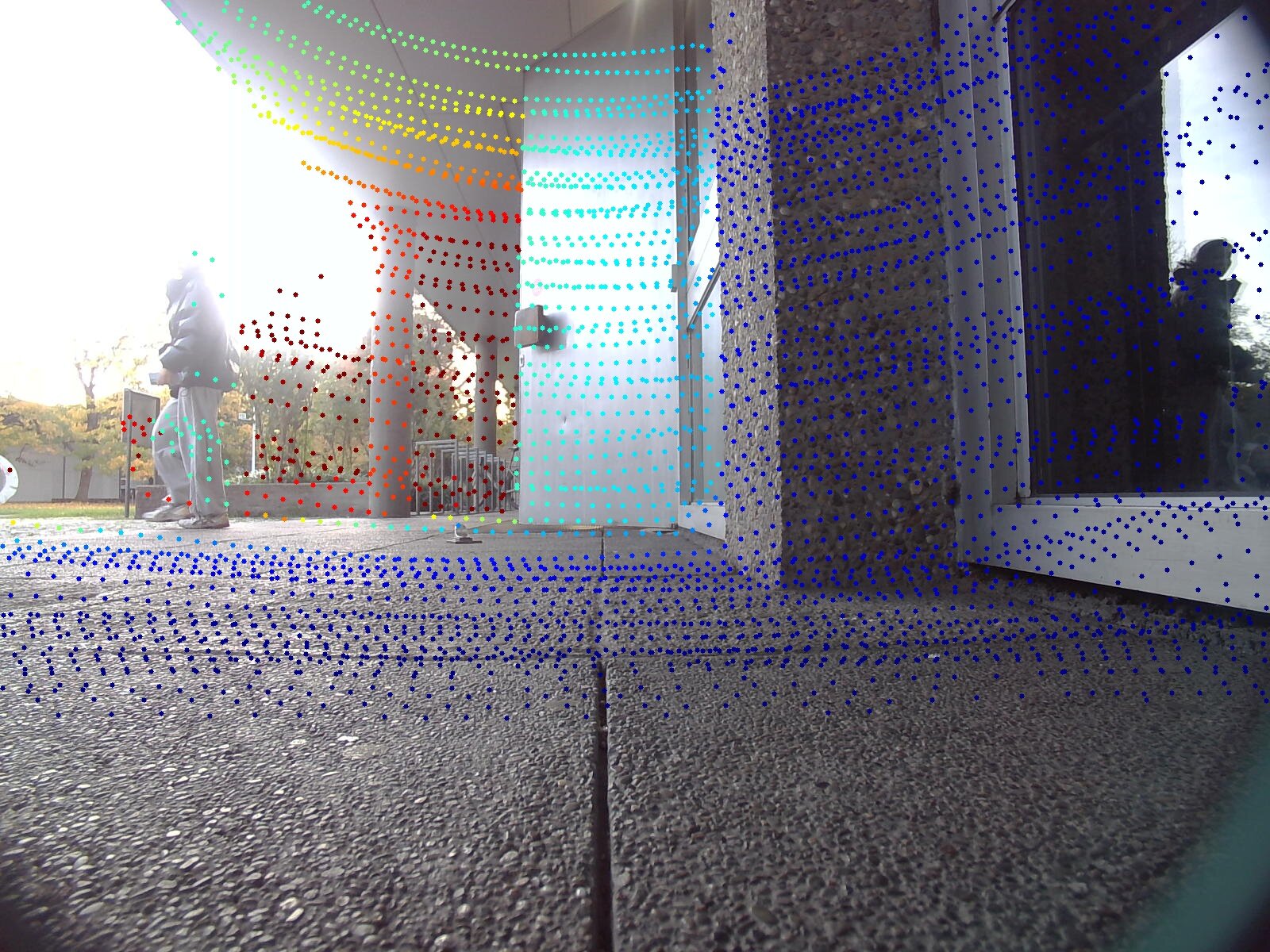} &
    \includegraphics[width=0.49\linewidth]{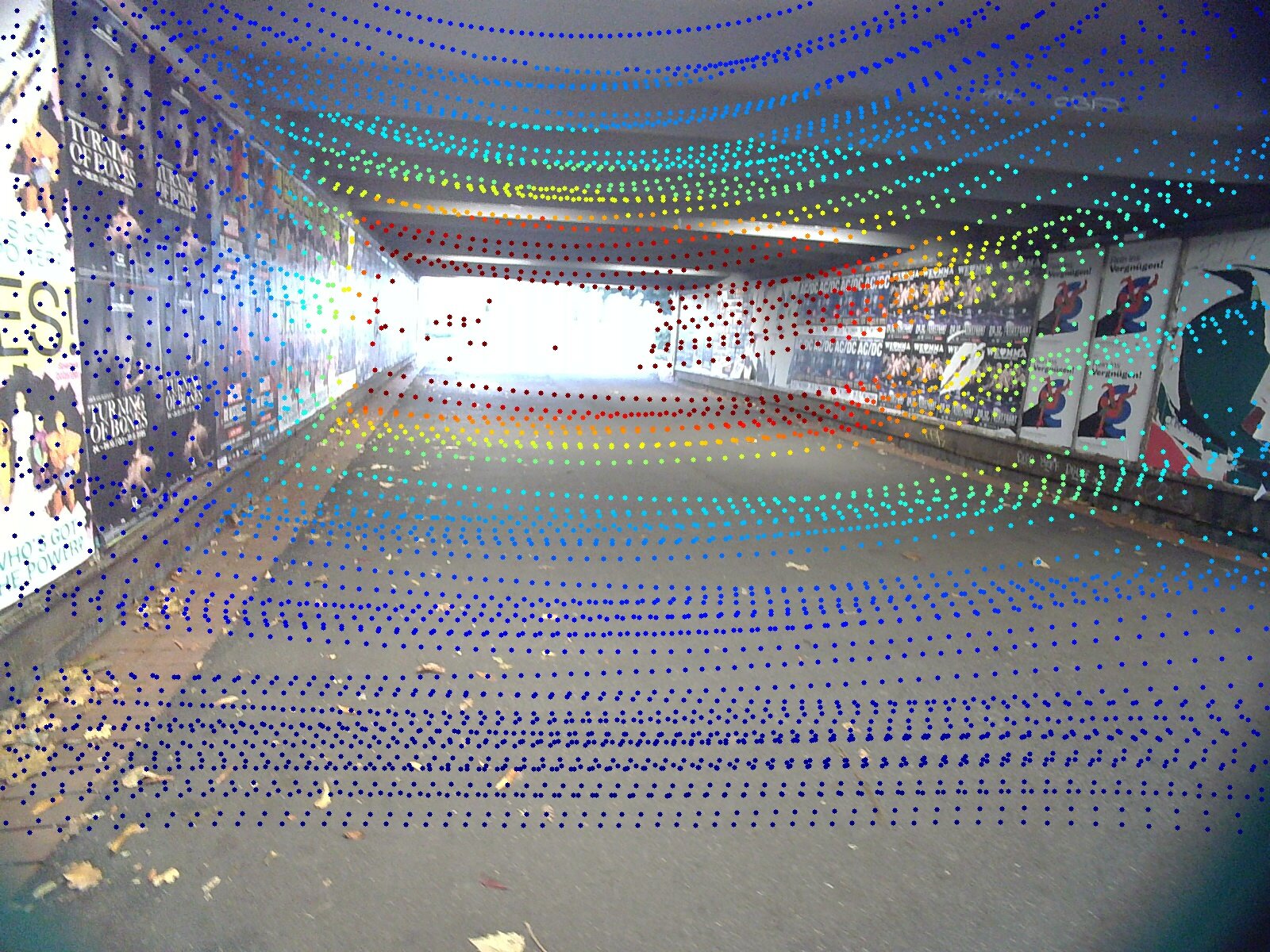} \\[0pt]
    \includegraphics[width=0.49\linewidth]{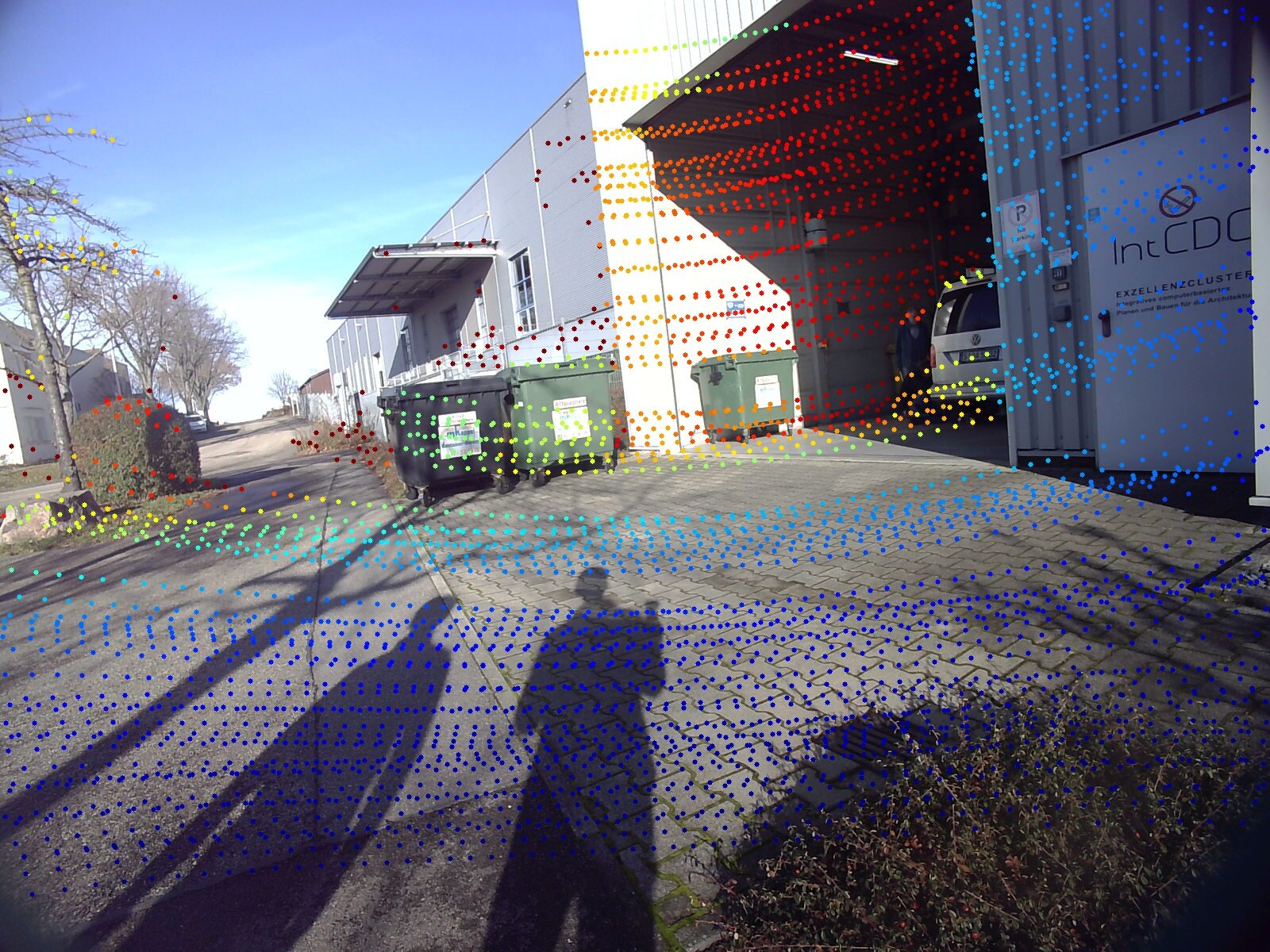} &
    \includegraphics[width=0.49\linewidth]{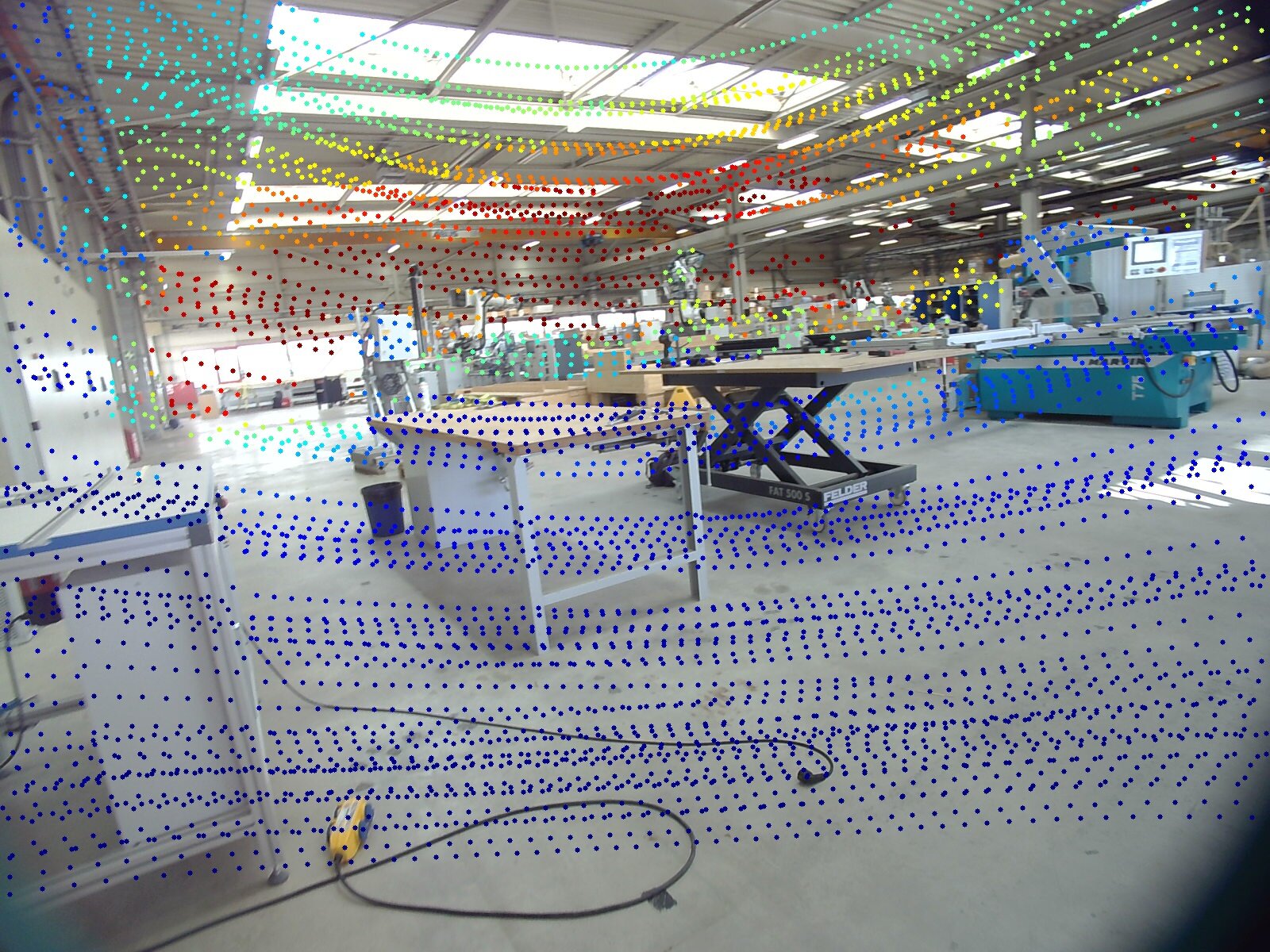} \\
\end{tabular}
\caption{Example camera images overlaid with projected LiDAR points, colorized by depth. \textbf{Top:} Stadtgarten scene near building (left) and the GNSS-denied underpass tunnel (right). \textbf{Bottom:} Construction Hall scene entrance area (left) and the indoor construction hall (right).}
\label{fig:scene_images}
\end{figure}    
\subsection{Stadtgarten Scene}
The Stadtgarten scene was collected in Stuttgart Stadtgarten, a public park with diverse GNSS visibility conditions. The operator walked through the environment at normal walking pace and held the device stationary over the checkpoints. Two sequences were captured covering the same area. Geodetic checkpoints are distributed across three distinct zones: open-sky areas, partially obstructed areas under tree cover or near building facades, and a GNSS-denied 30\,m-long underpass tunnel. As shown in Figure~\ref{fig:scene_images}, the tunnel interior is dark and has a highly symmetric cross-sectional geometry, which can cause LiDAR odometry to degenerate along the tunnel axis and makes this segment particularly challenging. Figure~\ref{fig:teaser} (top) illustrates the spatial distribution of the checkpoints.

\subsection{Construction Hall Scene}
The Construction Hall scene was collected at a large construction site (IntCDC, University of Stuttgart)\footnote{\url{https://www.intcdc.uni-stuttgart.de/research/research-infrastructure/}}. Two sequences were captured, each beginning and ending outdoors with good GNSS visibility and RTK fix, and traversing the interior of the construction hall where GNSS signals are severely degraded, with only weak receptions likely due to skylights or multipath reflections. The sequences cover the same area but with clockwise and counter-clockwise walking directions, providing complementary views of the outdoor-to-indoor transition. Table~\ref{tab:dataset} summarizes the statistics of the recorded sequences and Figure~\ref{fig:scene_images} shows representative camera images from the scene.

\subsection{Geodetic Ground Truth}
Accurate geodetic ground truth is a central contribution of this dataset, as it enables direct evaluation of absolute global positioning accuracy without reliance on SE(3) alignment. For both scenes, ground truth was established using a two-stage procedure. First, a set of open-sky anchor points was surveyed via static GNSS observations, achieving a position standard deviation of ${<}$5\,mm. A Leica TS16 total station was then oriented to these anchors and used to measure the remaining checkpoints via reflective prisms. These measurements covered points under GNSS obstruction or inside GNSS-denied areas, carrying the global reference frame into regions where kinematic GNSS is unavailable. For the Construction Hall scene, an additional instrument setup was required to reach the deeper indoor checkpoints. The survey was extended via a further traverse, where a second total station setup was established inside the building by resection onto targets measured from the outdoor station. This propagated the global reference frame through the full outdoor-to-indoor transition. The resulting ground truth covers both outdoor and indoor checkpoints with an estimated accuracy of better than 1\,cm. During data collection, the handheld device was carefully centered over these checkpoints and held stationary, enabling direct comparison between the estimated positions and the surveyed coordinates.

\subsection{Terminology}
Throughout this paper, \emph{GNSS} refers to the satellite navigation systems in general (GPS, Galileo, GLONASS, and BeiDou). \emph{RTK} refers specifically to the differential positioning mode in which the receiver resolves carrier-phase integer ambiguities using base-station corrections, achieving centimeter-level accuracy. When satellite signals or SAPOS corrections are unavailable or insufficient, the receiver falls back to RTK float, DGPS, or single-point positioning in order of decreasing accuracy. All of these modes are GNSS-based but not RTK-grade. We use ``GNSS-denied'' and ``GNSS-degraded'' to describe signal availability conditions, and ``GNSS quality'' or ``GNSS status'' to describe the positioning mode of the receiver. The surveyed reference points are called \emph{checkpoints} throughout this paper. In photogrammetric practice, ground control points (GCPs) are used to actively constrain a map during processing, whereas checkpoints are used solely to verify the accuracy of the final result. Since our surveyed points play no role in the estimation pipeline and serve purely as an independent accuracy reference, the term checkpoint is used.

\section{Experiments}
We first introduce the evaluated RTK-SLAM methods and the evaluation metrics, then present per-scene results for all four sequences, and conclude with a cross-cutting discussion of the key findings.

\subsection{Evaluated Methods}

\paragraph{FAST-LIO-SAM.}
FAST-LIO-SAM~\cite{wang2022fast_lio_sam} couples the FAST-LIO2~\cite{xu2022fast} LiDAR-inertial odometry front-end with the factor graph backend of LIO-SAM~\cite{shan2020lio}, with RTK measurements incorporated as loose-coupling GNSS position factors. We refer the reader to the respective papers for implementation details. For evaluation we report two types of results. \emph{Online} poses are estimated causally without access to future measurements. \emph{Offline} results are obtained by applying global pose graph optimization over the complete trajectory. The offline result is expected to yield higher accuracy, as batch optimization can propagate GNSS corrections across GNSS-denied sections in both directions.

\paragraph{OKVIS2-X.}
OKVIS2-X~\cite{boche2025okvis2} is a keyframe-based SLAM system with tightly coupled GNSS integration, supporting flexible sensor modality combinations within a unified factor graph. GNSS measurements are fused directly with visual reprojection and inertial factors by explicitly estimating a 4-DoF transformation $\mathbf{T}_{GW}$ that aligns the local SLAM frame with the global reference frame. We evaluate two configurations: \textbf{OKVIS2-X(lvig)} uses the full LiDAR-visual-inertial-GNSS modality set, while \textbf{OKVIS2-X(vig)} relies on visual-inertial-GNSS only. Both output a globally referenced trajectory for direct comparison with geodetic checkpoints, and the two configurations also serve as an ablation to quantify the contribution of LiDAR depth in GNSS-degraded conditions.

\begin{table*}[!ht]
\centering
\small
\caption{Quantitative accuracy results across all sequences and methods. We report absolute ATE for both online and offline results, as well as relative accuracy of the offline results after SE(3) alignment. Gap [\%] denotes the alignment-induced error reduction, revealing hidden systematic offsets or orientation errors. The best absolute ATE are marked in bold for each sequence.}
\label{tab:summary}
\adjustbox{max width=\linewidth,center}{%
\setlength{\tabcolsep}{4.5pt}
\begin{tabular}{@{}ll rrrr rrrr rrrr r@{}}
\toprule
 & & \multicolumn{4}{c}{\textbf{FAST-LIO-SAM}} & \multicolumn{4}{c}{\textbf{OKVIS2-X(vig)}} & \multicolumn{4}{c}{\textbf{OKVIS2-X(lvig)}} & \textbf{RTK} \\
\cmidrule(lr){3-6}\cmidrule(lr){7-10}\cmidrule(lr){11-14}
\textbf{Scene} & \textbf{Seq.} & Online & Offline & SE3 & Gap & Online & Offline & SE3 & Gap & Online & Offline & SE3 & Gap & Standalone \\
 & & [m] & [m] & [m] & [\%] & [m] & [m] & [m] & [\%] & [m] & [m] & [m] & [\%] & [m] \\
\midrule
Stadtgarten   & Seq.\,1 & 0.162 & \textbf{0.068} & 0.065 &  4 & 3.276 & 0.189 & 0.185 &  2 & 4.103 & \textbf{0.068} & 0.060 & 12 & 13.98 \\
Stadtgarten   & Seq.\,2 & 0.150 & 0.099 & 0.077 & 22 & 2.695 & 0.907 & 0.831 &  8 & 3.180 & \textbf{0.092} & 0.080 & 13 & 11.99 \\
Constr.\ Hall & Seq.\,1 & 0.256 & \textbf{0.248} & 0.220 & 11 & 1.437 & 0.788 & 0.579 & 27 & 0.761 & 0.321 & 0.227 & 29 & 12.01 \\
Constr.\ Hall & Seq.\,2 & 0.439 & 0.373 & 0.089 & 76 & 3.715 & 0.700 & 0.511 & 27 & 0.825 & \textbf{0.170} & 0.081 & 52 & 14.84 \\
\bottomrule
\end{tabular}}
\end{table*}

\subsection{Evaluation Protocol}
\label{sec:eval_protocol}

\paragraph{Coordinate transformation pipeline.}
RTK-SLAM methods estimate poses in a local east-north-up (ENU) Cartesian frame whose origin is fixed at the first valid RTK fix. Poses are expressed in the IMU body frame $I$. To compare against surveyed checkpoints, we first compute the 3D position of the device base center $B$ in the ENU world frame:
\begin{equation}
    \mathbf{p}_{W,B}^{(k)} = \mathbf{p}_{W,I}^{(k)} + \mathbf{R}_{W,I}^{(k)}\,\mathbf{t}_{I \to B},
    \label{eq:base_center}
\end{equation}
where $\mathbf{p}_{W,I}^{(k)}$ and $\mathbf{R}_{W,I}^{(k)}$ are the estimated position and orientation at timestep $k$ in the IMU body frame, and $\mathbf{t}_{I \to B}$ is the fixed offset from the IMU origin to the base center expressed in the IMU body frame (see Table~\ref{tab:calibration}). Since we focus on evaluating the positioning error, this reduces to position only. The resulting ENU position $\mathbf{p}_{W,B}^{(k)} = (e,\,n,\,u)^\top$ of the base center is then converted to geographic coordinates $(\varphi,\,\lambda,\,h)$. The ENU origin is the geodetic coordinates $(\varphi_0,\,\lambda_0,\,h_0)$ of the first valid RTK fix. Finally, $(\varphi,\,\lambda,\,h)$ is projected to UTM Zone~32U, all within the ETRS89 reference frame, which is the native coordinate system of the SAPOS correction service. The resulting easting, northing, and height $(\hat{E}_i,\,\hat{N}_i,\,\hat{h}_i)$ can be compared directly to the surveyed checkpoint coordinates $(E_i^*,\,N_i^*,\,h_i^*)$.

\paragraph{Absolute accuracy metric.}
Let $\boldsymbol{\varepsilon}_i = (\hat{E}_i - E_i^*,\;\hat{N}_i - N_i^*,\;\hat{h}_i - h_i^*)^\top$ denote the 3D position error at checkpoint $i$. The root-mean-square error over $N$ checkpoints is
\begin{equation}
    \mathrm{RMSE} = \sqrt{\frac{1}{N}\sum_{i=1}^{N} \|\boldsymbol{\varepsilon}_i\|^2}.
    \label{eq:rmse}
\end{equation}
This metric is a direct adaptation of the Absolute Trajectory Error (ATE)~\cite{sturm2012benchmark} to a globally referenced setting. The key difference from standard ATE is that no SE(3) alignment is applied. Since the estimated trajectory already resides in the target geodetic coordinate frame after the transformation described above, both global offset and accumulated drift can be reflected in the error. The timestamp of each checkpoint visit is pre-determined by detecting the stationary periods in the SLAM trajectory and matching them to the surveyed checkpoint locations. The resulting lookup table is then shared across all evaluated methods to ensure a consistent and reproducible evaluation.

\begin{figure*}[htbp]
    \centering
    \includegraphics[width=\linewidth]{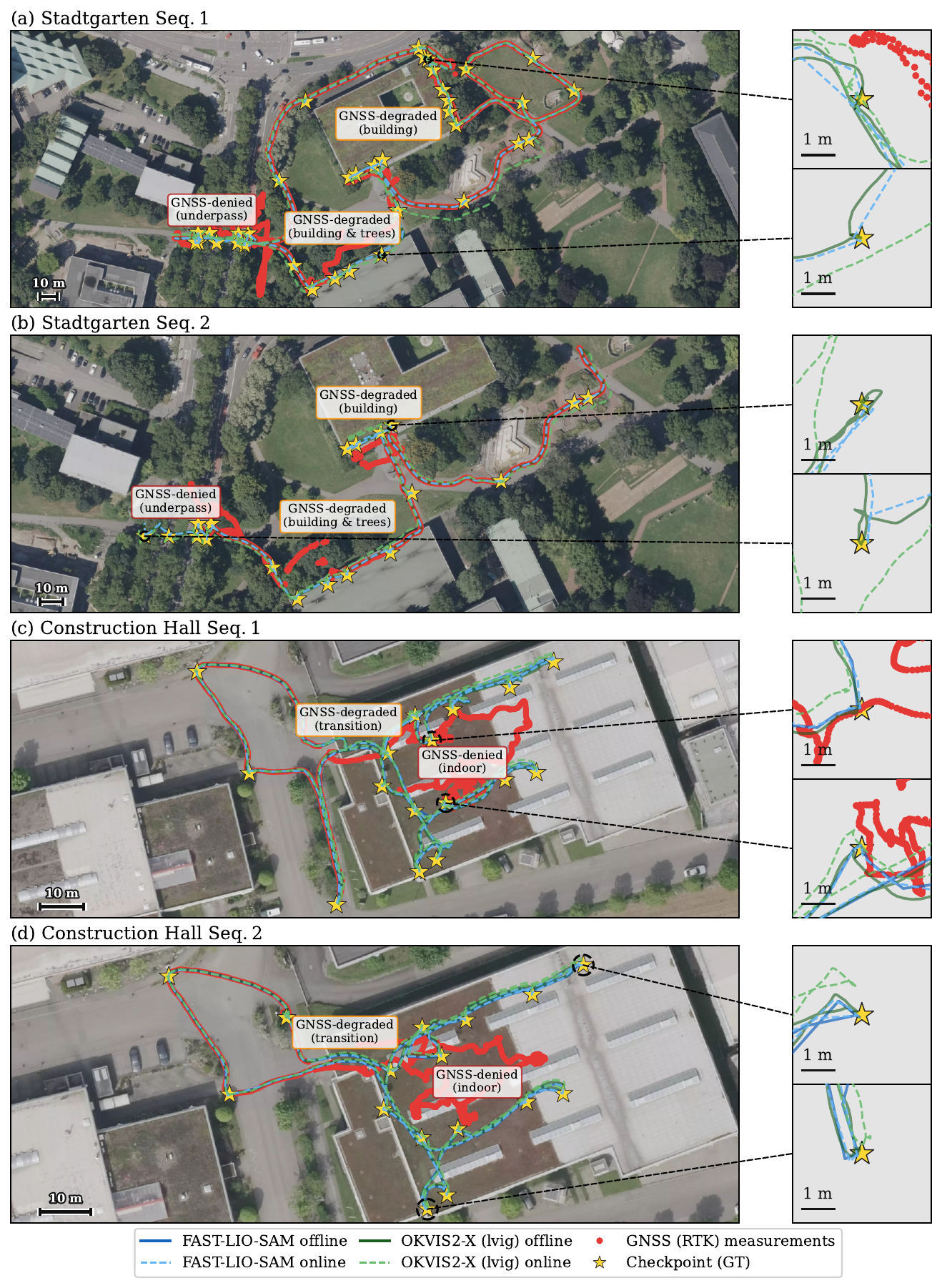}
    \caption{Trajectory comparisons for all four sequences on satellite imagery. Rows (a)--(b): Stadtgarten Seq.\,1--2; rows (c)--(d): Construction Hall Seq.\,1--2. GNSS-degraded zones are annotated. In the Construction Hall indoor environment, the GNSS receiver still obtains occasional measurements, but only at single-point positioning accuracy.}
    \label{fig:traj_all}
    \end{figure*}
\paragraph{SE(3)-aligned relative accuracy and the gap to absolute accuracy.}
To make the distinction between global and relative accuracy, we additionally compute for each method the SE(3)-aligned ATE. Given matched position pairs $\{(\hat{\mathbf{p}}_i, \mathbf{p}_i^*)\}$, we find the optimal rigid transformation
\begin{equation}
    (\mathbf{R}^*, \mathbf{t}^*) = \operatorname*{arg\,min}_{R \in \mathrm{SO}(3),\, \mathbf{t}} \sum_{i=1}^{N} \|\mathbf{R}\hat{\mathbf{p}}_i + \mathbf{t} - \mathbf{p}_i^*\|^2
    \label{eq:se3align}
\end{equation}
and report the RMSE of the residuals $\|\mathbf{R}^*\hat{\mathbf{p}}_i + \mathbf{t}^* - \mathbf{p}_i^*\|$. This is equivalent to the standard ATE used in SLAM benchmarking~\cite{sturm2012benchmark}. The alignment absorbs any constant systematic offset and global orientation error between the estimated and reference frame. The difference between the global absolute RMSE~(Eq.~\ref{eq:rmse}) and the SE(3)-aligned RMSE therefore quantifies how much of the absolute error is hidden by this alignment. For a RTK-SLAM system that successfully maintains global accuracy, these two values should be nearly the same. A large gap indicates that the system has significant global drift or systematic errors that the standard metric would miss.

\subsection{Results and Discussion}

We show the per-point errors and trajectories for all four sequences in Figure~\ref{fig:teaser} (bottom) and Figure~\ref{fig:traj_all}, with quantitative results summarised in Table~\ref{tab:summary}. Three distinct error regimes emerge as GNSS availability degrades. In open-sky zones, all RTK-SLAM methods achieve sub-5\,cm absolute accuracy. As GNSS degrades near building facades and trees, standalone RTK errors grow to several meters, while LiDAR-aided RTK-SLAM methods remain mostly within 10\,cm. In fully GNSS-denied zones, standalone RTK errors reach 10--40\,m, while the offline results of FAST-LIO-SAM and OKVIS2-X(lvig) maintain decimeter level accuracy by relying on LiDAR-inertial odometry to bridge the outage. The Construction Hall sequences represent a more challenging scenario. The GNSS-denied interior spans over 400\,s and 150\,m of travel, with only short outdoor segments at the start and end. Compared to Stadtgarten, absolute errors are substantially larger even in offline results, directly reflecting the longer GNSS outage.

\paragraph{Online vs. Offline Estimation}

Online results accumulate larger errors during GNSS outages, confirming the benefit of offline global optimization with entire observations, which propagates corrections both forward and backward in time. For the online results, especially for OKVIS2-X, significantly larger errors are observed. A possible cause is its global reference frame initialization strategy. It fixes the global orientation once the yaw uncertainty of the estimated 4-DoF world-to-global transformation falls below a fixed threshold (0.1°), which prevents further refinement when additional RTK fixes become available. In comparison, FAST-LIO-SAM continuously performs global batch optimization directly in the global frame.

\paragraph{LiDAR-aided vs. Vision-only}

Both LiDAR-aided methods achieve comparable absolute ATE of \textless 10\,cm in Stadtgarten. Without LiDAR, OKVIS2-X(vig) achieves 18.9\,cm in Seq.\,1 but degrades severely in Seq.\,2 (90.7\,cm) due to front-end divergence in the texture-poor underpass that LiDAR depth would otherwise prevent. In Construction Hall, OKVIS2-X(vig) performs substantially worse in absolute terms. Without direct LiDAR geometric observations to constrain the trajectory during the indoor section, visual-inertial drift accumulates more rapidly, and GNSS re-acquisition upon exiting the building can only partially recover the global position.

\paragraph{Absolute vs. Relative Accuracy}

The alignment gap in Stadtgarten is small for all methods (2--22\,\%), indicating that RTK effectively anchors the trajectories globally in this predominantly open-sky scene. In Construction Hall, gaps reach up to 76\,\%, suggesting that the brief outdoor sections with RTK fix are insufficient to precisely anchor the global coordinate frame. Beyond positional drift, a global systematic error can accumulate during the indoor traversal that SE(3) alignment absorbs but geodetic evaluation exposes.

\paragraph{SLAM-only vs.\ RTK-SLAM}
\begin{table}[h]
\centering
\small
\caption{Comparison of SLAM-only relative ATE vs.\ RTK-SLAM relative and absolute ATE.}
\label{tab:slam_only}
\setlength{\tabcolsep}{4pt}
\begin{tabular}{@{}lrrr@{}}
\toprule
\textbf{Seq.} & \makecell[r]{\textbf{SLAM-only}\\\textbf{Rel. [cm]}} & \makecell[r]{\textbf{RTK-SLAM}\\\textbf{Rel. [cm]}} & \makecell[r]{\textbf{RTK-SLAM}\\\textbf{Abs. [cm]}} \\
\midrule
Stadtgarten\,1 & 61.5 & \textbf{6.5} &  6.8 \\
Stadtgarten\,2 & 26.4 & \textbf{7.7} &  9.9 \\
Construction\,1 & 31.4 & \textbf{22.0} & 24.8 \\
Construction\,2 & 12.2 & \textbf{8.9}  & 37.3 \\
\bottomrule
\end{tabular}
\end{table}
Beyond providing a global coordinate reference, we analyze whether RTK integration can also improve relative trajectory accuracy. To quantify this, we evaluate the FAST-LIO-SAM result without RTK as a measure of SLAM-only relative accuracy, and compare it against its RTK-SLAM offline results in Table~\ref{tab:slam_only}. In both scenes, the relative accuracy of SLAM-only is worse than that of RTK-SLAM, indicating the positive impact of RTK integration in correcting long-range odometry drift, particularly in sequences without loop closure opportunities.

\begin{figure}[htbp]
\centering
\includegraphics[width=1.01\linewidth]{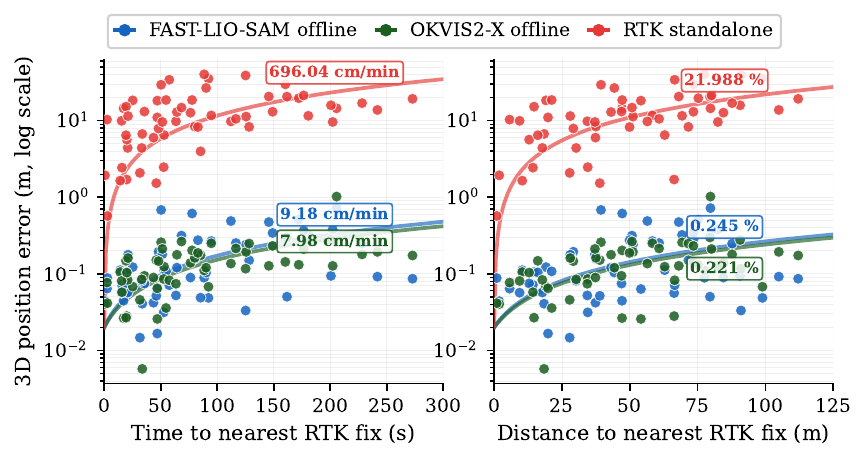}
\caption{3D absolute positioning error (log scale) as a function of time (left) and distance (right) to the nearest RTK fix, aggregated over the measurements from all sequences. Regression lines indicate the average drift rate.}
\label{fig:rtk_corr}
\end{figure}

\paragraph{Drift Behavior Under GNSS Outage}

We focus this analysis on the offline results, which represent the best accuracy from each system. Figure~\ref{fig:rtk_corr} shows how absolute error grows with GNSS outage duration and travel distance. Because offline optimization propagates RTK corrections both forward and backward, the x-axis is the distance to the nearest RTK fix, rather than only since the last fix. The y-axis is shown on a logarithmic scale to visualize the wide error range across methods. A linear drift model $\epsilon(x) = \epsilon_0 + \alpha x$ is fitted with the intercept fixed at $\epsilon_0 = 2$\,cm, representing the standard error of a typical RTK fix measurement. The slope $\alpha$ captures the drift that accumulates during the outage. A clear positive correlation can be observed across all methods, confirming that error grows steadily with proximity to RTK coverage. The key observation is that RTK-SLAM methods maintain effective position tracking even as GNSS quality degrades. The fitted drift rates are low: {9.2\,cm/min} (0.25\,\% of path length) for FAST-LIO-SAM and {8.0\,cm/min} (0.22\,\%) for OKVIS2-X(lvig), reflecting the effectiveness of LiDAR odometry in bounding dead-reckoning drift during GNSS outages. In contrast, standalone RTK degrades significantly once signal quality deteriorates. These drift rates can explain the absolute ATE results of the Construction Hall sequences.

\section{Conclusion}
We have presented a RTK-SLAM dataset and evaluation methodology that together address a critical gap in the field. The dataset provides synchronized LiDAR, camera, IMU, and RTK inputs alongside geodetic ground truth established independently via total station and static GNSS, enabling direct evaluation of absolute global positioning accuracy. To complement the dataset, we propose an evaluation protocol based on direct global accuracy metrics, explicitly addressing the limitation of SE(3)-aligned ATE, which absorbs global drift and can underestimate absolute errors by up to 76\%. We evaluate different sensor-fusion configurations and observe that LiDAR-aided systems achieve consistently higher absolute accuracy, and offline results obtained through global optimization outperform online estimates, suggesting that surveyors can meaningfully improve output quality with a post-processing step. When GNSS degrades, standalone RTK deteriorates rapidly to meter-level errors, whereas RTK-SLAM methods maintain decimeter-level absolute accuracy even through GNSS-denied indoor environments. Future work will investigate tight fusion of raw GNSS measurements into the RTK-SLAM estimator, which we expect to further improve global accuracy during GNSS-degraded transitions.

\section{Acknowledgements}
Supported by the Deutsche Forschungsgemeinschaft (DFG, German Research Foundation) under Germany´s Excellence Strategy – EXC 2120/1 –  390831618.

{
	\begin{spacing}{1.17}
		\normalsize
		\bibliography{isprs_slam_rtk} 
	\end{spacing}
}

\end{document}